\documentclass[11pt]{article}
\usepackage[T1]{fontenc}        %
\usepackage{dialogue}           %
\usepackage{tabularx}           %
\usepackage{amssymb}            %
\usepackage{amsmath}            %
\usepackage{algorithm}          %
\usepackage{algpseudocode}      %
\usepackage{float}              %
\usepackage{mdframed}           %
\usepackage{xcolor}             %
\usepackage{threeparttable}     %
\usepackage{graphicx}           %
\usepackage{enumitem}           %
\usepackage{authblk}            %
\usepackage{url}

\title{Representation Learning to Study Temporal Dynamics in Tutorial Scaffolding}

\author{Conrad Borchers\textsuperscript{1*}}
\author{Jiayi Zhang\textsuperscript{2}}
\author{Ashish Gurung\textsuperscript{1}}

\affil{\textsuperscript{1}Carnegie Mellon University}
\affil{\textsuperscript{2}Worcester Polytechnic Institute}
\affil{\texttt{*cborcher@cs.cmu.edu}}

\date{} %

\begin{document}

\maketitle

\begin{abstract}
Adaptive scaffolding enhances learning, yet the field lacks robust methods for measuring it within authentic tutoring dialogue. This gap has become more pressing with the rise of remote human tutoring and large language model-based systems. We introduce an embedding-based approach that analyzes scaffolding dynamics by aligning the semantics of dialogue turns, problem statements, and correct solutions. Specifically, we operationalize alignment by computing cosine similarity between tutor and student contributions and task-relevant content. We apply this framework to 1,576 real-world mathematics tutoring dialogues from the Eedi Question Anchored Tutoring Dialogues dataset. The analysis reveals systematic differences in task alignment and distinct temporal patterns in how participants ground their contributions in problem and solution content. Further, mixed-effects models show that role-specific semantic alignment predicts tutorial progression beyond baseline features such as message order and length. Tutor contributions exhibited stronger grounding in problem content early in interactions. In contrast, student solution alignment was modestly positively associated with progression. These findings support scaffolding as a continuous, role-sensitive process grounded in task semantics. By capturing role-specific alignment over time, this approach provides a principled method for analyzing instructional dialogue and evaluating conversational tutoring systems.
\end{abstract}

\section{Introduction}

Scaffolding refers to the dynamic adjustment of instructional support in response to learners' evolving needs. Understanding scaffolding dynamics is critical for explaining instructional effectiveness and informing the design of tutoring systems. Prior research emphasizes that this effectiveness arises from the bidirectional nature of tutoring dialogue, in which learner responses continuously shape subsequent feedback and scaffolding. This interactive structure enables well-designed tutoring systems to approximate the effectiveness of human tutors \cite{vanlehn2011relative}.

Scaffolding remains difficult to operationalize in naturalistic tutorial dialogue. Prior research in AIED has typically relied on rule-based representations of instructional support. For example, dialog-based systems such as AutoTutor provide scaffolding through predefined tutorial dialogue moves, including prompts, hints, and summaries, while using latent semantic analysis to assess student contributions against expert-authored expectation texts \cite{graesser2000teaching}. In contrast, step-based systems operationalize scaffolding through structured hint sequences and quantitative measures of hint usage level, such as assistance scores \cite{long2013supporting}. While these approaches have been influential, they impose categorical distinctions on inherently continuous instructional phenomena and are often tied to specific tasks or system architectures. As conversational tutoring increasingly occurs in open-ended, text-based settings (e.g., remote tutoring \cite{gurung2025human} or through large language models (LLMs) \cite{schmucker2024ruffle}), there is a growing need for task-grounded, scalable measures of scaffolding that generalize across task contexts.

We propose an embedding-based framework that models scaffolding as a continuous semantic property of tutorial dialogue. We characterize tutor and student turns by their alignment with two task anchors: the problem statement and the correct solution. Alignment with the problem reflects dialogue that remains close to the task formulation, whereas alignment with the solution captures the extent to which support becomes more directly oriented toward the target answer. Together, these dimensions define a continuum of scaffolding from problem-focused reasoning to directive, solution-oriented support, consistent with prior work on multi-level hinting that progresses from reflective prompts to explicit solution disclosure \cite{aleven2016help}. We apply this approach to real-world tutoring data to examine role differences in alignment and test whether these measures predict instructional progression beyond message order and length. Accordingly, we ask: \textbf{RQ1}: How do tutor and student roles differ in their semantic alignment to the problem statement and the correct solution during problem solving? \textbf{RQ2}: How does semantic alignment between tutorial dialog and problem statement or solution \textit{evolve} during problem solving? \textbf{RQ3}: Do role-specific patterns of semantic alignment predict problem progression beyond message order and length?

This work makes three contributions. First, we introduce a continuous operationalization of scaffolding based on semantic alignment to task content. Second, we show that embedding-based similarity captures role-specific and temporal patterns of instructional support in natural dialogue. Third, we demonstrate that these measures predict instructional progression beyond turn-based measures. Together, these contributions advance task-grounded methods for studying scaffolding in both human and conversational tutoring systems.

\section{Related Work}

\subsection{Scaffolding in Natural Language Dialogue for Learning}

Conversational tutoring systems support learning through natural language interaction that resembles human tutoring \cite{graesser2000teaching,schmucker2024ruffle}. Dialogical systems prompt learners to articulate their reasoning and engage in mixed-initiative exchanges with the tutor. This shift from action-based input in step-based tutoring systems to language-based interaction enables richer modeling of student understanding and has been shown to support learning across domains and educational levels \cite{paladines2020systematic}.

A prominent example is AutoTutor, which models tutoring as a sequence of predefined dialogue moves, including questions, prompts, hints, and summaries \cite{graesser2000teaching,graesser2005autotutor}. AutoTutor assesses learning by analyzing the semantic content of students' natural language responses. Using latent semantic analysis (LSA), the system compares student input to expert-authored expectation texts that represent ideal answer components. Conceptual coverage is quantified through cosine similarity between these representations, with predefined thresholds determining whether specific expectations have been satisfied. These similarity estimates are used to update a student model and guide subsequent instructional moves.

This architecture enables adaptive scaffolding within dialogue. When student responses are incomplete or vague, the system prompts elaboration and provides progressively specific hints to guide learners toward target concepts without directly revealing answers \cite{graesser1995collaborative,graesser2000teaching}. However, these approaches rely on expert-authored expectation texts, manually defined similarity thresholds, and task-specific representations of knowledge. Such dependencies limit scalability and constrain applicability in open-ended tutoring contexts, including remote human tutoring and emerging LLM-based systems \cite{gurung2025human,schmucker2024ruffle}.

\subsection{Advances in Analyzing Tutorial Dialogue and Educational Text}

LSA represents meaning through distributional word co-occurrence but remains static and limited in capturing context-dependent semantics. In contrast, LLM-based sentence embeddings provide representations with large context windows \cite{reimers-2019-sentence-bert}. Unlike LSA's fixed semantic space derived from expert corpora, these embeddings leverage general-purpose models to produce richer and more adaptable representations of learning discourse \cite{borchers2025disentangling}. This shift supports more generalizable approaches to modeling language in educational settings.

Accordingly, LLMs have been increasingly integrated into digital learning systems, including tools such as Khanmigo and MathTutor \cite{shetye2024evaluation,venugopalan2024combining}. Recent work has applied LLMs to support problem solving, estimate student knowledge, and generate adaptive feedback aligned with established tutoring principles \cite{schmucker2024ruffle,scarlatos2025exploring,venugopalan2024combining}. Beyond direct tutoring, sentence embeddings have also been used to analyze instructional dialogue, including identifying effective tutoring strategies and modeling students' conceptual understanding \cite{lin2022good,borchers2025disentangling}. This body of work motivates the hypothesis that embedding-based semantic similarity can operationalize instructional scaffolding as a continuous property of tutor and learner behavior during problem solving, which we investigate in this study.

\section{Methods}

\subsection{Data Set}

We analyzed the Eedi-2K ``Question-Anchored Tutoring Dialogues'' dataset, an open-source corpus containing real-world, chat-based math tutoring interventions \cite{zent2025piivot}. Eedi is an online platform that provides students with practice problems aligned with school mathematics curricula and facilitates one-to-one chat-based tutoring during problem solving. The dialogues in this dataset are anchored to individual mathematics problems, with each interaction centered on helping a student work through a specific problem. From this corpus, we analyzed 1,576 complete dialogues, totaling 55,322 conversational moves. The dataset included 25 unique human tutors across several students and no student IDs.

\subsection{Data Preprocessing and Feature Engineering}
\label{sec:methods:preproc}

We used sentence embeddings to transform dialogue into high-dimensional vector representations, enabling quantitative analysis of semantic relationships. Specifically, we used the \texttt{all-MiniLM-L6-v2} transformer model from the Sentence Transformers library \cite{reimers-2019-sentence-bert} to generate 384-dimensional semantic embeddings for all textual content. We encoded three textual components: (1) individual tutor and student dialogue messages, (2) the diagnostic math problem text, and (3) the correct solution text. We then computed cosine similarity between sentence embedding vectors to operationalize key scaffolding dimensions: the semantic alignment of dialogue moves with the problem statement, alignment with the correct solution, and the baseline task structure between question and solution. 

We examined two temporal features as baselines: (1) absolute message sequence rank within a dialogue, and (2) relative normalized position within the dialogue. For each dialogue, we computed relative position as $n/N$, where $n$ is the message's sequence index and $N$ is the total number of messages. Relative position is operationalized as the progression of dialogue toward problem completion (0 = start, 1 = end). This normalization by dialogue length allows progression to be compared across interactions.

\subsection{Investigating Temporal Scaffolding Dynamics (RQ1 \& RQ2)}
\label{sec:methods:rq2}

To answer \textbf{RQ1}, we compared (i) the distributions of alignment scores by role using density histograms and (ii) the role-specific temporal trajectories described above. Together, these distributional and temporal comparisons show how tutors and students align with problem and solution content during tutoring. To answer \textbf{RQ2}, we quantified how semantic alignment to the problem statement and the reference solution changes over time. We computed cosine similarity between each dialogue turn and (i) the problem text and (ii) the solution text, then summarized these scores as a function of dialogue progress. Specifically, we normalized each turn's position within its dialogue to $n/N$ (0=start, 1=end), averaged alignment scores across all turns at each relative position, and applied Gaussian smoothing to estimate the overall temporal trajectories.

\subsection{Predicting Progression from Semantic Alignment (RQ3)}

To answer \textbf{RQ3}, we tested whether semantic alignment to the problem statement and reference solution predicts \emph{relative dialogue progression} beyond non-semantic baselines. Our outcome variable was dialog progress (see Section \ref{sec:methods:rq2}). We fit a series of linear mixed-effects regression models while logit-transforming progression scores to account for the bounded outcome. All continuous predictors were standardized (mean = 0, SD = 1) for interpretability. All models included tutor random intercepts to account for clustering of messages within tutors. We also checked model assumptions, finding no notable multicollinearity (all VIFs < 1.44), approximately normally distributed residuals (skewness = 0.03, kurtosis = 0.33), and no evidence of heteroscedasticity.

We established a baseline model (Model~0) using only absolute message sequence rank, capturing the assumption that dialogue progresses monotonically as turns accumulate. Model~1 added message length as a predictor. Model~2 introduced cosine similarity to the problem statement and to the reference solution, testing whether semantic alignment explains variation in progression beyond message order and length. Model~3 further allowed these semantic effects to vary by tutor and student. Model comparisons were based on the Bayesian Information Criterion (BIC) and likelihood ratio tests. Supplemental analysis code to reproduce our analyses is available online \cite{repo}.

\section{Results}

Descriptively, tutors contributed 59.4\% of messages across an average of 65.6 problems. Dialogue sessions averaged 35.1 messages (median = 31; range = 20--148). Messages were short ($M$ = 29.3 characters; median = 21.0; SD = 29.1).

\subsection{Temporal Alignment Trends (RQ1 \& RQ2)}

\begin{figure}[H]
    \centering
    \includegraphics[width=\textwidth]{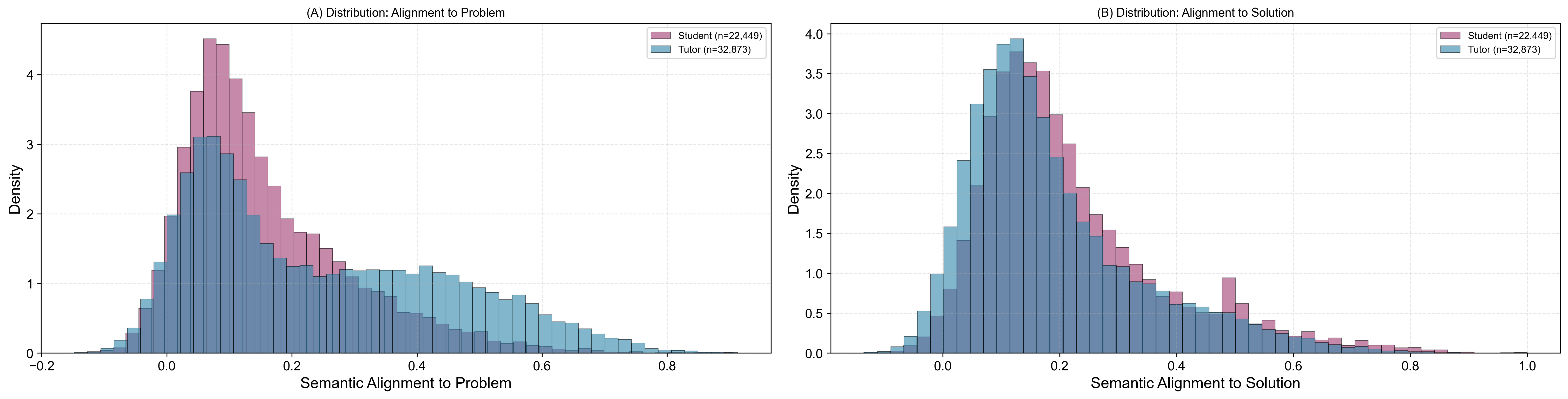}
    \caption{
        Density distributions of semantic alignment scores for tutor and student moves. 
        \textbf{(A)} Alignment to the problem statement.
        \textbf{(B)} Alignment to the correct solution.
    }
    \label{fig:distributions}
\end{figure}

We begin by examining the distributions of semantic alignment in dialogues (RQ1). Figure~\ref{fig:distributions} presents the density distributions of alignment scores for tutor and student moves. The distribution of tutor alignment to the problem statement is bimodal (Figure~\ref{fig:distributions}A). This suggests that tutors alternate between two instructional modes: one more aligned with the problem content (peak around 0.4) and another, more conversational or exploratory (peak around 0.1). In contrast, the student moves cluster exclusively at lower alignment values, indicating limited convergence toward the problem language. For solution alignment (Figure~\ref{fig:distributions}B), both distributions are left-skewed with minimal tutor-student differentiation.

\begin{figure}[H]
    \centering
    \includegraphics[width=\textwidth]{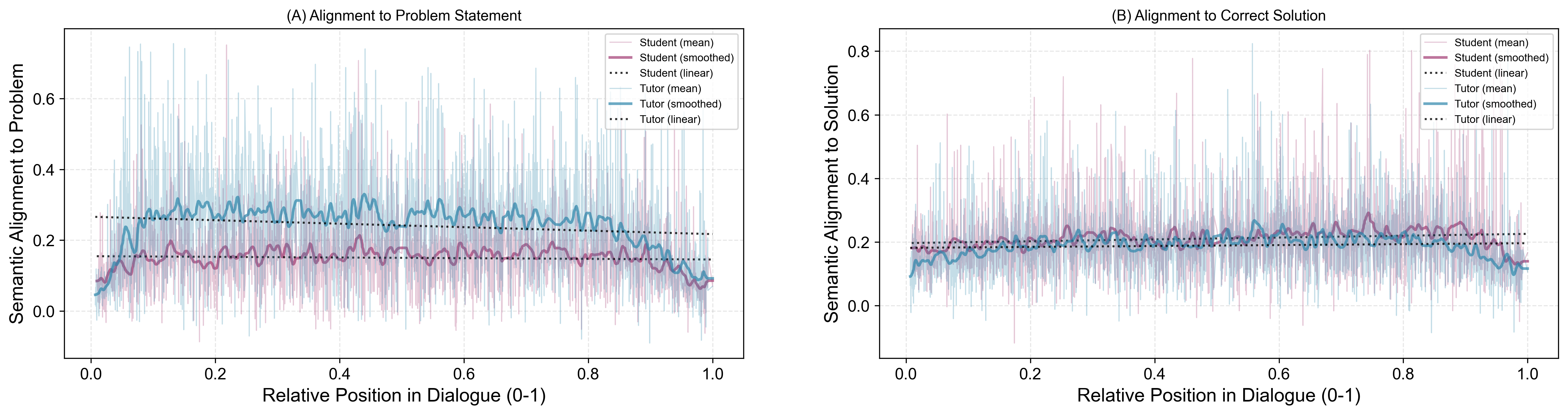}
    \caption{
        Temporal evolution of semantic alignment.
        \textbf{(A)} Alignment to the problem statement.
        \textbf{(B)} Alignment to the correct solution.
    }
    \label{fig:temporal-alignment}
\end{figure}

The temporal analysis for RQ2 (Figure~\ref{fig:temporal-alignment}) demonstrates that tutor alignment to the problem statement decreases (after initially rising) as the normalized position progresses from 0 to 1. In contrast, student alignment to the problem remains consistently low. Both roles show minimal alignment to the solution throughout dialogues. These temporal patterns suggest tutors and students maintain problem focus while gradually increasing solution-related content.

\subsection{Predicting Progression from Semantic Alignment (RQ3)}

We examined whether semantic similarity measures predicted progression (the \% of total dialogue completed at a given dialogue move; see Section \ref{sec:methods:preproc}) through tutoring dialogues using four nested linear mixed models with logit-transformed progression scores, with tutor random effects to account for across-tutor variance. All predictors were standardized. Adding message length significantly improved fit over sequence alone, $\chi^2(1) = 78.24$, $p < .001$, $\text{BIC} = 264500$. Adding cosine similarity to question and solution provided further improvement, $\chi^2(2) = 28.29$, $p < .001$, $\text{BIC} = 264494$. Allowing these effects to differ by role yielded additional improvement, $\chi^2(2) = 32.43$, $p < .001$, with the full model achieving $\text{BIC} = 264483$, indicating the best balance of fit and parsimony.

For the chosen model, tutor random effects were small (ICC = .01). Sequence position was the strongest positive predictor of dialog progression ($\beta = 1.72$, $p < .001$), with a smaller positive effect of message length ($\beta = 0.14$, $p < .001$). Question similarity (tutor: $\beta = -0.04$, $p = .004$; student: $\beta = -0.09$, $p < .001$) and tutor solution similarity ($\beta = -0.07$, $p < .001$) were negatively associated with progression, whereas student solution similarity was positively associated ($\beta = 0.07$, $p < .001$). These results indicate that problem-focused alignment occurs earlier in dialogues, while solution-focused alignment emerges later, particularly in student contributions (controlling for all other predictors).

\section{Discussion}

This study examined an embedding-based approach to modeling instructional scaffolding in tutorial dialogue as a continuous, semantically grounded process. The results demonstrate the effectiveness of embedding-based methods for evaluating student–tutor interactions, extending their use in AIED beyond prior applications such as knowledge tracing~\cite{scarlatos2025exploring}.

\subsection{Semantic Alignment With the Problem}

The distribution of semantic alignment between dialogue turns and the problem statement sheds light on how tutors construct scaffolding during interaction (RQ1). Student turns cluster around low alignment, indicating minimal reuse of problem language. In contrast, tutor turns exhibit a bimodal pattern, with a dominant low-alignment mode and a smaller peak near 40\%. The former may reflect dialogue focused on prompting and feedback, while the latter captures moments when tutors restate or paraphrase key problem elements to establish instructional grounding~\cite{clark1991grounding}. This pattern is consistent with prior accounts of scaffolding, in which tutors primarily guide learners through questioning while selectively restating task constraints to clarify goals and reduce cognitive load~\cite{chi2001learning}.

Temporal analyses (RQ2) show that tutors continue to occasionally reference problem content as solutions develop. This suggests that restatement is deployed opportunistically to diagnose understanding and consolidate progress, consistent with micro-adaptive models of tutoring~\cite{katz2014summarization}. Overall, these findings characterize scaffolding as a continuous, flexible process grounded in task semantics.

\subsection{Across-Turn Semantic Alignment With the Solution}

Solution alignment exhibits a distinct pattern from problem alignment. Both tutor and student turns follow overlapping, unimodal distributions, and alignment to the reference solution remains largely stable over time. This pattern is consistent with effective scaffolding that supports learners in constructing solutions while avoiding conditions associated with the ``fluency heuristic''~\cite{chi2001learning}, whereby smooth interaction is mistaken for understanding. If tutors frequently revealed answers or students primarily sought confirmation, one would expect higher and increasing solution alignment, along with dialogue that promotes shallow learning and inflates perceived proficiency through misleading performance cues~\cite{carpenter2020students}.

\subsection{Predicting Progression Using Semantic Alignment}

Using mixed-effects regression models (RQ3), we examined how the semantic similarity of tutor and student utterances to the problem statement and reference solution relates to dialogue progression, while adjusting for message order and length. Students and tutors exhibited distinct trends. Both referenced problem content more frequently early in the dialogue. In contrast, their solution similarity increased over time, suggesting greater use of solution-aligned language as they approach task completion, consistent with task progression.

Although statistically reliable, these effects are modest in magnitude. Nevertheless, the results support cosine similarity as a meaningful proxy for further study of scaffolding. Embedding-based measures enable fine-grained analysis of instructional dynamics beyond traditional metrics such as assistance scores~\cite{aleven2016help}, and provide a foundation for studying both human and LLM-based tutoring.

\subsection{Limitations and Future Work}

Scaffolding occurs both within problems, as studied here, and across problems through processes such as fading. The absence of student identifiers prevents analysis of longitudinal scaffolding and its optimality for learning. We leave validation against learning outcomes and systematic evaluation of potentially excessive LLM scaffolding to future work~\cite{borchers2026large}. Additionally, incorporating alternative semantic anchors could capture forms of scaffolding beyond the problem-solution spectrum, including metacognitive support and multiple reasoning paths, perhaps in a direct comparison with LSA-based approaches \cite{graesser2000teaching,graesser2005autotutor}.

Several extensions can broaden the scope of our framework. First, scaffold episodes could be identified by detecting bursts of high tutor problem alignment and testing whether these precede increases in student solution alignment, providing evidence for targeted restatement of task constraints~\cite{chi2001learning,katz2014summarization}. Second, clustering tutors by alignment trajectories may reveal distinct scaffolding profiles and clarify whether observed bimodality reflects stable strategies or adaptive behavior, extending prior work on dialogue-based support~\cite{graesser2000teaching,katz2014summarization}.

\section{Conclusion}

Despite scaffolding's central role in tutoring, existing approaches provide limited means for analyzing how instructional support unfolds in natural dialogue. We introduce an embedding-based framework that models scaffolding as continuous semantic alignment between dialogue, problems, and solutions. This approach captures interpretable, role-specific dynamics in tutorial interaction. Tutors predominantly ground early exchanges through alignment with problem content, while students increasingly adopt solution-aligned language as they progress, controlling for move sequence. This transition reflects a gradual transfer of responsibility from tutor to learner. More broadly, these findings demonstrate the value of embedding-based representations for analyzing instructional dialog.

\bibliographystyle{splncs04}
\bibliography{main} %

\end{document}